\pdfoutput=1
\documentclass[journal]{IEEEtran}
\usepackage[table]{xcolor}
\usepackage{amsmath,amsfonts}
\usepackage{algorithmic}
\usepackage{algorithm}
\usepackage{array}
\usepackage[caption=false,font=normalsize,labelfont=sf,textfont=sf]{subfig}
\usepackage{textcomp}
\usepackage{stfloats}
\usepackage{url}
\usepackage{verbatim}
\usepackage{graphicx}
\usepackage{cite}
\usepackage{multirow}
\usepackage{pgfplots}
\pgfplotsset{compat=1.18}
\usepackage{tikz}
\begin{document}

\title{
Spatial Hierarchy and Temporal Attention Guided Cross Masking for Self-supervised Skeleton-based Action Recognition
}

\author{Xinpeng Yin, Wenming Cao,~\IEEEmembership{Senior Member,~IEEE}, 
        \thanks{This work is supported by the Shenzhen Science and Technology Program under Grant JCYJ20220531100814033 and the National Natural Science Foundation of China under grant 61771322. Corresponding author: Wenming Cao}
        \thanks{Wenming Cao and Xinpeng Yin are with the State Key Laboratory of Radio Frequency Heterogeneous Integration (College of Electronics and Information Engineering, Shenzhen University) (e-mail: wmcao@szu.edu.cn; 2110436215@email.szu.edu.cn)}
        }

\markboth{Journal of \LaTeX\ Class Files,~Vol.~14, No.~8, August~2021}%
{Shell \MakeLowercase{\textit{et al.}}: Hyperbolic Hierarchy-Guided Spatiotemporal Cross Masking for Unsupervised Skeleton-Based Action Recognition}



\maketitle

\begin{abstract}
In self-supervised skeleton-based action recognition, the mask reconstruction paradigm is gaining interest in enhancing model refinement and robustness through effective masking.
However, previous works primarily relied on a single masking criterion, resulting in the model overfitting specific features and overlooking other effective information.
In this paper, we introduce a hierarchy and attention guided cross-masking framework (HA-CM) that applies masking to skeleton sequences from both spatial and temporal perspectives.
Specifically, in spatial graphs, we utilize hyperbolic space to maintain joint distinctions and effectively preserve the hierarchical structure of high-dimensional skeletons, employing joint hierarchy as the masking criterion.
In temporal flows, we substitute traditional distance metrics with the global attention of joints for masking, addressing the convergence of distances in high-dimensional space and the lack of a global perspective.
Additionally, we incorporate cross-contrast loss based on the cross-masking framework into the loss function to enhance the model's learning of instance-level features.
HA-CM shows efficiency and universality on three public large-scale datasets, NTU-60, NTU-120, and PKU-MMD. 
The source code of our HA-CM is available at https://github.com/YinxPeng/HA-CM-main.
\end{abstract}


\section{Introduction}\label{sec: intro}
\IEEEPARstart{A}Human action recognition has consistently posed a significant challenge in computer vision. It finds extensive applications such as human-computer interaction\cite{rIN1}, medical rehabilitation\cite{rIN2}, and video surveillance\cite{rIN3}. 
Since 3D skeletal data offers advantages over RGB, optical flow, and depth information—such as greater computational efficiency\cite{rIN4}, resilience to background noise\cite{rIN5}, and enhanced privacy\cite{rIN6} protection—skeleton-based action recognition has gained significant attention. To further alleviate the reliance on labeled data, self-supervised learning for skeleton-based recognition offers a compelling alternative to traditional supervised methods.

Self-supervised skeleton-based action learning primarily involves two paradigms: contrastive learning\cite{r1, r2, r3} and mask reconstruction\cite{r11, r12, r13}. Contrastive learning forms positive and negative sample pairs to optimize their representation distance, enabling the model to learn invariant semantics and discriminative features. However, it relies heavily on heuristic data augmentation and often boosts performance by increasing the volume of contrastive pairs, resulting in a bloated model. In contrast, mask reconstruction uses an encoder-decoder structure to mask and reconstruct parts of the data, encouraging the model to learn representations of these masked features, resulting in a more streamlined approach. The method proposed in this paper is also based on the mask reconstruction paradigm.

Previous works \cite{r12, r13} have demonstrated that masking joints with detailed information compel the model to learn masked edge features, enhancing its information extraction capabilities. To ensure generalization, introducing randomness during masking is essential. However, these methods mainly focus on the temporal aspects of skeleton sequences, which can lead to overfitting. For instance, masking based solely on motion intensity can cause the model to overemphasize high-intensity movements while neglecting details in low-intensity or static poses. To address this bias, this paper explores masking strategies for skeleton sequences from both spatial and temporal perspectives.

From a temporal perspective, pivotal research \cite{r12} calculates the Euclidean distance of the same joint between adjacent frames using its 3D coordinates, treating this distance as the joint's motion intensity at a given moment to mask joints in high-dimensional space.
However, relying on low-dimensional relationships for high-dimensional masking does not adequately reflect the diversity and complexity of joints. 
Moreover, as the dimension $n$ increases, the relative distances between joints, calculated as \(d(x, y) = \sqrt{\sum_{i=1}^n (x_i - y_i)^2}\), tend to converge, diminishing the effectiveness of distance information. 
Consequently, Euclidean distance calculations often remain confined to low-dimensional space, hindering a deeper understanding of motion patterns. This approach also captures only local motion information and lacks a global regulatory perspective.

To tackle these challenges, this paper employs the sum of the inner products, referred to as attention, to establish a more effective masking criterion in the temporal flow. 
The inner product calculation is fundamentally linear, relying on the relationships of angles and magnitudes between vectors, which makes it less susceptible to the challenges posed by increasing dimensionality. 
Note that angle relationships tend to converge in high-dimensional space. However, joints are arranged in an Euclidean chain in the temporal flow, indicating linear connections.
This linearity better reflects the actual movement of the joints. Overemphasizing angular changes can complicate the model's understanding of motion patterns and potentially mislead analyses. Therefore, reducing the emphasis on angular differences in high-dimensional space during temporal flow is essential. Additionally, the inner product calculation considers both local and global perspectives, facilitating a more comprehensive understanding of the temporal flow.

Shifting to spatial considerations, the skeleton is structured as a non-Euclidean hierarchical tree, which introduces additional complexities. The distances and angles between joints, as well as their hierarchical relationships, are crucial.
Utilizing the inner product in this context may overlook significant details, leading to an incomplete understanding of motion patterns. 
Thus, there is an urgent need for a method that preserves both the distinctions between joints and the hierarchical structure of the skeleton in high-dimensional space. 
This paper introduces hyperbolic space to address this challenge.

The unique negative curvature of hyperbolic space effectively preserves the relative positional relationships between key joints in high-dimensional data, mitigating the information loss associated with distance concentration in Euclidean space. Moreover, as a conformal transformation, hyperbolic mapping maintains the consistency of angles between joints, ensuring accurate motion analysis. 
Additionally, the exponential growth characteristic of hyperbolic space effectively represents the hierarchical structure of the skeleton\cite{RIN7}, enhancing the model's understanding of skeletal sequences. This also provides solid support for skeleton masking based on the hierarchy of joints.

In temporal flow and spatial graphs, joints with high correlation and low hierarchy often retain more detailed information. Therefore, we selectively mask these joints to compel the model to learn this edge information.
To enhance the coupling between spatial and temporal masks, we propose a hierarchy \& attention guided cross-masking (\textbf{HA-CM}) framework. 
Specifically, based on the temporal redundancy of the skeleton sequence, we cross-segment it into two parts using odd and even indexing, applying distinct masking strategies from spatial and temporal perspectives.
The masked features are then reassembled based on their original indices as input for reconstruction, facilitating interaction between joint features during reconstruction and enhancing the model's ability to perceive different types of information.

The cross-masking framework divides the masked features into two parts. Despite variations in detailed information due to different masking methods, they maintain high similarity at the instance-level, as subsets of the same sample. 
Thus, in designing the loss function, we guide the model to learn both the reconstruction loss of intra-sample details and introduce a cross-contrast loss ($\mathcal{L}_{c^2}$) to enhance the encoder's learning of instance-level features.

 The main contributions of this work can be summarized as:
\begin{itemize}
\item{We propose a hierarchy \& attention guided cross-masking (HA-CM) framework that applies masking to skeleton sequences from spatial and temporal perspectives, mitigating biases inherent in single masking methods.}
\item{We specifically utilize hyperbolic space to maintain joint distinctions and effectively preserve the hierarchical structure of high-dimensional skeletons, using the joint hierarchy as the criteria for masking.}
\item{In designing the loss function, we incorporated cross-contrast loss to enhance the model's capacity for learning instance-level features. Meanwhile, we validated the effectiveness of our proposed HA-SM across three large-scale datasets.}
\end{itemize}

\section{Related Work}
\subsection{Self-supervised Skeleton Representation Learning}
Self-supervised skeleton representation learning can be broadly categorized into encoder-decoder models and contrastive learning methods. The former focuses on learning joint features within individual samples, while the latter examines instance features across multiple samples.

In contrastive learning, Rao et al.\cite{r1} used eight data augmentation techniques, including rotation and shearing, to generate diverse positive and negative samples, highlighting the importance of augmentation in enhancing representation quality.
Guo et al. 
\cite{r2} enhanced feature generalization with extreme augmentations, while Zhang et al. \cite{r3} proposed a gradual augmentation strategy to generate ordered positive pairs, fostering consistent learning from different perspectives. These invariant contrastive methods ensure consistency in feature representations before and after transformations, which may sometimes result in the loss of critical information. To address this, Lin et al. \cite{r4} developed an equivariant learning network to better handle feature changes with transformations, thereby boosting performance.
Additionally, contrastive learning naturally aligns with multi-stream networks. Li et al. \cite{r5} introduced a cross-stream knowledge mining strategy to exploit complementary information across modalities. In contrast, Mao et al. \cite{r6} employed bidirectional knowledge distillation to improve cross-modal information transfer. HiCo \cite{r7} utilized multi-level features and performs hierarchical contrastive learning.
Hu et al. \cite{r42} introduced a Global and Local Contrastive Learning framework to leverage similarities between global and local crops of the same skeleton sequence to improve semantic learning and generalization performance.

For encoder-decoder models, LongT GAN \cite{r8} applied an autoencoder with adversarial training to reconstruct corrupted sequences. P\&C \cite{r9} further refined the decoder to enhance encoder learning.
\cite{r14} utilizes point cloud technology to color joints with both coarse and fine granularity and learns spatial-temporal features using a dual-stream autoencoder.
Recently, He et al. \cite{r10} proposed a masked reconstruction approach, utilizing low information density in images to reconstruct original signals. Although skeleton sequences are dense compared to RGB images and optical flow data, recognizing actions often requires only partial joint movements. Building on this, SkeletonMAE \cite{r11} used masked reconstruction at joint and frame levels, while MAMP \cite{r12} masked high-motion regions to emphasize edge features. From the perspective of choosing reconstruction targets, Xu et al. \cite{r13} employed a teacher-student model to generate advanced latent features.
Zhu et al.\cite{r48}
integrated masked skeleton feature reconstruction with a visual-language pre-trained model to leverage the strengths of both modalities. 
These masking strategies typically focus on temporal aspects and overlook spatial hierarchies. This paper explores how spatially-informed masking, leveraging the hierarchical nature of joints and hyperbolic, impacts the quality of learned representations.

\subsection{Hyperbolic Feature Embedding}
Since Ganea et al.\cite{r15} pioneered hyperNNs, introducing hyperbolic equivalents for fully connected layers and logistic regression in Euclidean, hyperbolic has gained significant attention. 
Subsequent advancements include hyperbolic convolutional neural networks \cite{r16}, hyperbolic graph neural networks \cite{r17}, and hyperbolic attention networks \cite{r18}. Compared to Euclidean, hyperbolic naturally excels at handling hierarchical and tree-like structures and representing complex high-dimensional data in lower dimensions, making it an attractive choice in deep learning. 
For instance, in gesture recognition, Leng et al.\cite{r19} proposed a dynamic hyperbolic attention network that leverages hyperbolic growth to better preserve mesh geometry and enhance feature differentiation based on similarity. In general face anti-spoofing, Hu et al.\cite{r20} introduced a novel hierarchical prototype-guided distribution refinement framework, which learns embedded features in hyperbolic to improve hierarchical relationship construction.
In information retrieval, Yan et al. \cite{r32} utilized hyperbolic to enhance the hierarchical structure between events and event types, alleviating the issue of vocabulary mismatch.

In supervised skeleton representation learning, Peng et al.\cite{r21} reconstructed GCNs on Riemannian manifolds, reducing model sizes by 60$\%$ while maintaining performance compared to dynamic graph generation methods. In self-supervised learning, Franco et al.\cite{r22} combined hyperbolic mapping with self-paced learning within a contrastive learning framework. They matched online, and target view features through hyperbolic mapping and used hyperbolic uncertainty to guide the learning process.
Chen et al. \cite{r23} embedded network outputs into hyperbolic and used a multi-layer perceptron (MLP) to convert the module into a homotopy mapping, enhancing supervisory signals and capturing the high-dimensional nonlinear structure of skeleton sequences.
However, these self-supervised action representation methods focus on capturing complex nonlinear relationships at the instance level, overlooking the commonality between hyperbolic and the hierarchical nature of skeletal graphs. In contrast, this paper combines hyperbolic with another self-supervised paradigm, masked reconstruction, forcing the model to learn fine-grained features within skeleton sequences.

\section{Preliminaries}\label{Sec3_1: preliminaries}
\subsection{Notations}
We outline the key notations, including representations for skeleton sequences, the Poincaré ball model, and hypergraph structures.
\begin{itemize}
    \item \textbf{Skeleton sequences:} Skeleton sequences are represented as \( \textbf{X} \in \mathbb{R}^{L \times J \times C} \), where \( L \) is the number of frames, and \( J \) is the number of joints. Each frame is a graph \( \mathcal{G} = (\nu, \varepsilon) \), where \( \nu = \{v_1, v_2, \cdots, v_J\} \) is the set of joints, and \( \varepsilon \) represents their topological relationships. The channel dimension \( C \) initially has a value of 3, representing the 3D coordinates \((x, y, z)\) of each joint in Euclidean space.
    
    \item \textbf{Poincaré ball model:} The Poincaré ball model \( (\mathbb{P}^n, g^{\mathbb{P}}) \) is defined as \( \mathbb{P}^n = \{ v \in \mathbb{R}^n \mid \|v\|^2 < -\frac{1}{c} \} \), where \( \mathbb{P}^n \) is the open unit ball in \( \mathbb{R}^n \). Here, \( v \) is a point in \( \mathbb{R}^n \), \( \|v\| \) is its L2 norm, and \( c \) (\( c < 0 \)) is a constant related to the curvature. The Riemannian metric \( g^{\mathbb{P}} = \left(\lambda_v^{c}\right)^2 g^{\mathbb{E}}\) is conformal to the Euclidean metric \( g^{\mathbb{E}} \) with the conformal factor \( \lambda_v^c = \frac{2}{1 - \|v\|^2} \).

\end{itemize}

\subsection{Hyperbolic Learning}
Due to the unique geometric and topological properties of the Poincaré ball model, vectors require the introduction of gyrovector \cite{r33} for calculations. For instance, the addition of two vectors \( u, v \in \mathbb{P}^n \), known as Möbius addition, is defined as:
\begin{equation}
u \oplus v = \frac{(1 + 2c \langle u, v \rangle + c\|v\|^2)u + (1 - c\|u\|^2)v}{1 + 2c \langle u, v \rangle + c^2 \|u\|^2 \|v\|^2}
\label{eq:mobius_addition}
\end{equation}
where \( \langle u, v \rangle \) represents the Euclidean inner product, and \( \|u\| \) and \( \|v\| \) denote the L2 norms of \( u \) and \( v \). The distance between two points \( u, v \in \mathbb{P}^n \) in the Poincaré ball model is given by the Poincaré distance formula:
\begin{equation}
d_{\mathbb{P}}(u, v) = \frac{2}{\sqrt{-c}} \tanh^{-1} \left(\sqrt{-c} \| - u \oplus v \|\right)
\label{eq:poincare_distance}
\end{equation}
where \( \| - u \oplus v \|\) is the L2 norm of the vector resulting from Möbius addition.

\section{Method}\label{Sec: method}

\begin{figure*}[!t]
\centering
\includegraphics[width=7in]{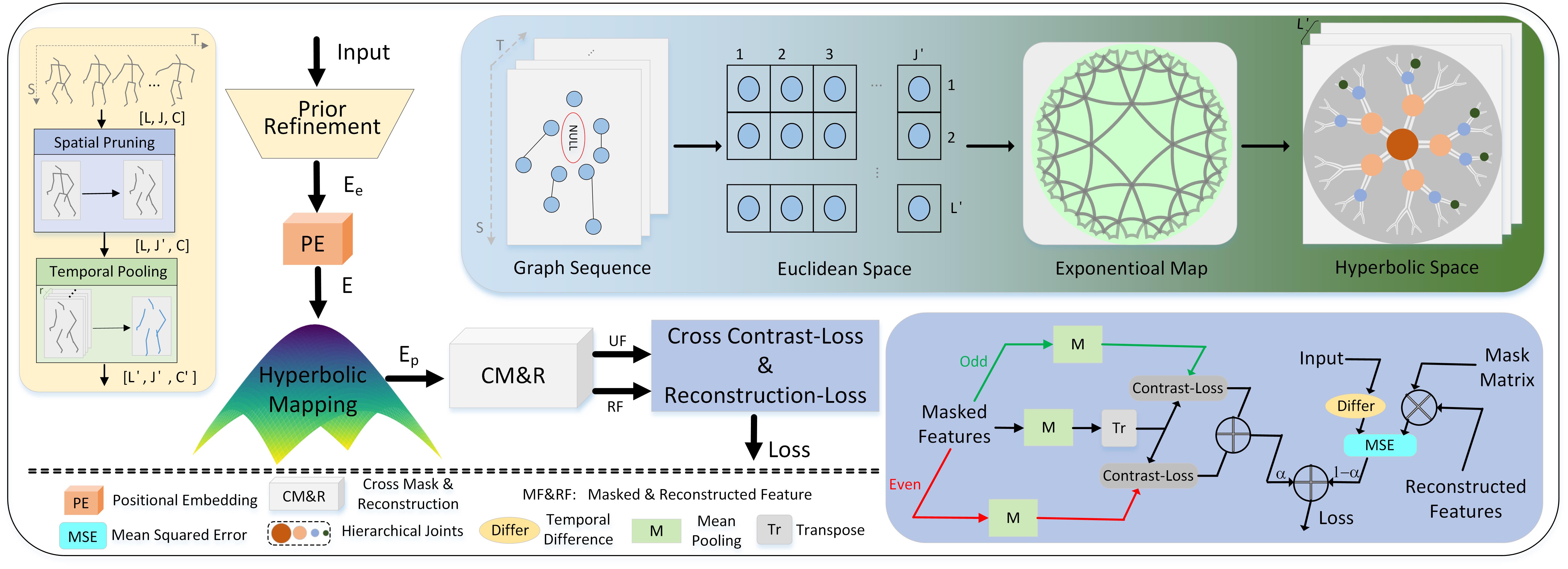}
\caption{(Match components with colors)\textbf{Architecture Overview of HA-CM.} 
The symbols $\textbf{E}_{e}$, $\textbf{E}$ and $\textbf{E}_{\mathbb{P}}$ correspond to those used in the main text. Note that the entire sequence is embedded in a high-dimensional hyperbolic space along the spatial dimension, while the computation of the masking criteria that determines which joints in different components should be masked is performed in a different space.}
\label{fig2}
\end{figure*}

\subsection{Pipeline Overview}
Figure \ref{fig2} illustrates the overall pipeline of the proposed
\underline{H}ierarchy \& \underline{A}ttention guided  \underline{C}ross \underline{M}asking (HA-CM) framework. 
To begin with, leveraging the spatiotemporal redundancy in skeleton sequences, we design a prior refinement module, including spatial pruning ($S_p$) and temporal pooling ($T_p$). This module reduces the number of redundant input tokens by eliminating joints located in the torso and applying local pooling to adjacent frames, thereby reducing model inference time while maintaining effectiveness. After refining the features, we apply positional embedding and then transform the joint features from Euclidean space to hyperbolic space using exponential mapping. This transformation more effectively captures the hierarchical structure of the skeleton graph.

Next, the mapped features are fed into the \underline{C}ross \underline{M}ask $\&$ 
\underline{R}econstruction (CM$\&$R) module, where they undergo masking and reconstruction, resulting in both the encoder's masked features and the decoder's reconstructed features, as shown in Figure \ref{Fig3}. 
Finally, we calculate the contrastive loss between the masked features and their odd and even components to guide the model in learning instance-level features across samples. Simultaneously, we compute the mean squared error (MSE) loss between the reconstructed features and the original sequence's motion information to constrain the model's learning of intra-sample detailed features. The final loss is obtained by weighting these two losses.

After the aforementioned pre-training, besides the general Encoder, the prior refinement and hyperbolic mapping components are integrated into downstream applications.

\subsection{Prior Refinement}
To address temporal redundancy, the proposed approach \cite{r12} divides the skeleton sequence \( \textbf{X} \in \mathbb{R}^{L \times J \times C} \) into non-overlapping segments and pools joint features with the same spatial position within each segment, thereby reducing the number of input tokens.
Building on this, our work explores a prior processing method from a spatial perspective to achieve this reduction further.

During data analysis, we found that the joints in the torso of the skeleton sequence exhibit minimal movement, meaning they carry limited useful information. However, in the masked reconstruction paradigm, these low-information joints are often retained because they are considered to carry coarse-grained features. While retaining coarse-grained information can force the model to learn finer details, this is only effective if the retained joints provide meaningful guidance for model training.
Therefore, this retention may lead to wasted training resources and cause the model to overlook more representative joint information. To enhance the model's efficiency and accuracy, we chose to exclude the torso joints from the skeleton sequence:

\begin{equation}\label{eq:sp}
\textbf{X}^{\prime} = S_p(\textbf{X}) \in \mathbb{R}^{L \times J^{\prime} \times C}
\end{equation}
where \( \textbf{X} \cap \textbf{X}^{\prime} = \{V_i\} \), where \( i \) denotes the joint indices of the torso part. After pruning, the sequence undergoes convolutional pooling \( T_p \) to obtain the embedding feature \( \textbf{E}_{e} \):

\begin{equation}\label{eq: tp}
\textbf{E}_{e} = T_p\left(\textbf{X}^{\prime}\right) \in \mathbb{R}^{L^{\prime} \times J^{\prime} \times C^{\prime}}
\end{equation}
In \( T_p \), the convolution kernel size, stride, and segment length are all \( r \), resulting in \( L / L^{\prime} = r \), \( C^{\prime} \) represents the dimension of the embedding features.

\subsection{Positional Embedding} \label{sec: pe}
The positional embedding in this work follows the proposed approach \cite{r12}, the spatial positional embedding $\textbf{P}_{s}$ and  temporal positional embedding $\textbf{P}_{t}$ are added to the embedding feature \( \textbf{E}_{e} \):
\begin{equation}
\textbf{E}=\textbf{E}_{e}+\textbf{P}_{t}+\textbf{P}_{s}
\end{equation}
where $P_{s}\in\mathbb{R}^{1\times J^{\prime}\times C^{\prime}}$ and $P_{t} \in\mathbb{R}^{L^{\prime}\times 1\times C^{\prime}}$ indicates that the spatial position embedding for the same joint across different frames and the temporal position embedding for all joints within the same frame are consistent. Broadcasting is used here to ensure that the position embedding and \( \mathbf{E} \) have the same dimensionality.

\subsection{Hyperbolic Mapping} \label{sec: hm}
The skeleton graph is a hierarchical tree structure, where the hierarchy progressively diminishes from the trunk (parent nodes) to the extremities of the limbs (child nodes). Compared to Euclidean space, hyperbolic space, with its negative curvature, more effectively represents the hierarchical structure of the skeleton graph. 
Therefore, in this work, we first project the embedded features from Euclidean space to hyperbolic space using an exponential map. The transformation is defined as:

\begin{equation}\label{eq:exp}
\textbf{E}_{\mathbb{H}} = \tanh(\sqrt{-c} \cdot \|\textbf{E}\|) \cdot \frac{\textbf{E}}{\sqrt{-c} \cdot \|\textbf{E}\|}
\end{equation}
After the exponential mapping, we further constrain the representation by projecting it onto the Poincaré ball. This projection is given by:

\begin{equation}\label{eq:poin}
\textbf{E}_{\mathbb{P}} = \frac{\textbf{E}_{\mathbb{H}}}{1 + \frac{\|\textbf{E}_{\mathbb{H}}\|^2}{-c}}
\end{equation}
where \( c \) is the curvature of the hyperbolic space, and \( \|\textbf{E}\| \) and \( \|\textbf{E}_{\mathbb{H}}\| \) are the Euclidean and hyperbolic norms of \( \textbf{E} \) and \( \textbf{E}_{\mathbb{H}} \), respectively.

\subsection{Cross Mask $\&$ Reconstruction}
\begin{figure*}[!t]
\centering
\includegraphics[width=7in]{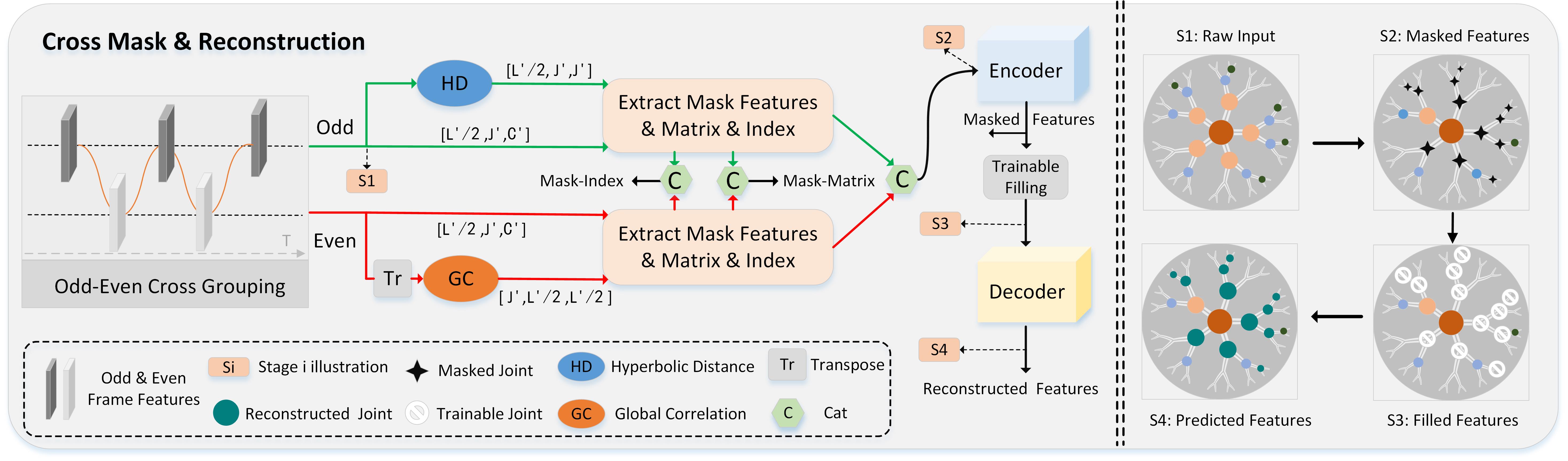}
\caption{The details of \textbf{CM\&R}.The green line corresponds to the spatial aspect, while the orange line represents strategy 1 of the temporal aspect.
The right side of the figure illustrates the masking and reconstruction process after the joints are embedded in hyperbolic space, incorporating randomness.}
\label{Fig3}
\end{figure*}

To address the limitations of overfitting specific types of features with a single masking approach, this work implements masking from both spatial and temporal perspectives on the skeleton sequences.
Additionally, to enhance the coupling between spatiotemporal masks, we construct a cross-mask $\&$ reconstruction framework that leverages the temporal redundancy of the skeleton sequences, as illustrated in Figure \ref{Fig3}.
The basic architecture is Encoder-Decoder and
before feeding features into the Encoder, the process includes three parts: odd-even cross-grouping, mask criteria calculation, and non-masked joint extraction.

\noindent\textbf{Odd-Even Cross-Grouping:} Based on the time dimension index, we divide the features into two parts: $\textbf{P}_o$ and $\textbf{P}_e$:
\begin{equation}\label{po}
C_g(\textbf{E}_{\mathbb{P}}) =
\left\{
\begin{aligned}
& \textbf{P}_o = \textbf{E}_{\mathbb{P}[i,:,:]}, \quad i \text{ is odd} \\
& \textbf{P}_e = \textbf{E}_{\mathbb{P}[i,:,:]}, \quad i \text{ is even}
\end{aligned}
\right.
\end{equation}

\noindent\textbf{Mask Criteria Calculation:}
Note that the function of the mask criterion is solely to determine the index positions of the joints fed into the Encoder, without affecting the joint features themselves. Therefore, the mask criterion can be computed based on the original input, the encoded features, or the embedded features after hyperbolic mapping.

In $\textbf{P}_o$, the masking criterion is based on the spatial hierarchy of the skeleton graph. 
In hyperbolic space, the hierarchical structure of nodes is typically characterized by both the radial distance from the root node and the hyperbolic distances between nodes. Radial distance indicates a node's absolute position within the hierarchy, with greater distances generally corresponding to lower hierarchical levels. Hyperbolic distance, on the other hand, reflects the relative positioning between nodes, where those farther from others are usually less central. In skeletal sequences, joints that are lower in the hierarchy or non-central often contain finer details. Thus, this paper assesses joint hierarchy using these metrics and applies them as the criteria for masking.
The mask criterion here requires the joints to reside in hyperbolic space.

The root node in skeletal sequences is generally assumed to belong to the torso region of the human body. In this paper, we designate the fused representation of the joints from the torso region, excluded in Section \ref{sec: hm}, as the root node. 
To ensure that the root node and the other joints remain in the same feature space, we apply the same temporal pooling operation described in Eq.\ref{eq: tp} to the root node and the remaining joints while sharing convolutional weights.
The features of the root node can be represented as:
\begin{equation}
\textbf{E}_{\text{root}} = \text{Meanpooling}\left(T_p\left(\textbf{X} \cap \textbf{X}^{\prime}\right)\right) \in \mathbb{R}^{L^{\prime} \times 1 \times C^{\prime}}
\label{eq:root}
\end{equation}
the mean pooling operation aims to aggregate the features of the torso joints in space, resulting in a single root node for each frame. 

After $\textbf{E}_{\text{root}}$ undergoes hyperbolic mapping described in Sec.\ref{sec: hm} to obtain $\textbf{E}_{\mathbb{P}{\text{root}}}$,
we utilize Eq.\ref{eq:poincare_distance} to compute the radial distances $\textbf{D}_{\text{radial}} \in \mathbb{R}^{L^{\prime} \times J^{\prime} \times 1}$ between all joints and the corresponding root node, and the radial distance of the 
$j_{th}$ joint at the 
$k_{th}$ frame is expressed as:
\begin{equation}
\textbf{D}_{\text{radial}, j}^k = d_{\mathbb{P}}\left(\textbf{E}_{\mathbb{P}{\text{root}}}^k, \textbf{E}_{\mathbb{P}{\text{j}}}^k\right)
\end{equation}
where $\textbf{E}_{root}^k$ and $\textbf{E}_{\mathbb{P}{\text{j}}}^k$ represent the root joint and the $j_{th}$ joint in the $k_{th}$ frame, respectively.

To calculate the hyperbolic distances between joints, we construct a hyperbolic distance matrix (\(\textbf{HDM} \in \mathbb{R}^{L^{\prime}/2 \times J^{\prime} \times J^{\prime}}\)) based on Eq.\ref{eq:poincare_distance}. The hyperbolic distance between the $i_{th}$ and $j_{th}$ joints in the $k_{th}$ frame is given by:
\begin{equation}
\textbf{HDM}^{k}_{i j}=d_{\mathbb{P}}\left(\textbf{E}_{\mathbb{P}{\text{i}}}^k, \textbf{E}_{\mathbb{P}{\text{j}}}^k\right)
\end{equation}

Finally, we concatenate the even-indexed part of the radial distance $\textbf{D}_{\text{radial}}$ and hyperbolic distance matrices \textbf{HDM} along the last dimension to obtain the complete hierarchical information matrix \textbf{HIM} representing the joints:
\begin{equation}
\textbf{HIM}=Concat(\textbf{D}_{\text{radial}}, \textbf{HDM})
\in \mathbb{R}^{L^{\prime}/2 \times J^\prime \times (J^\prime + 1)}
\end{equation}
each joint's hierarchical information is represented by \(J+1\) values, including a "0" value representing the hyperbolic distance between the joint and itself.
We sum these \(J+1\) distance values to obtain the final representation of the joint's hierarchy $\textbf{S}_{H}$, which also serves as the criteria for the joint's mask:
 \begin{equation}\label{eq:SH}
    \mathbf{S}_{H} = \sum_{i=0}^{J^{\prime}} \textbf{HIM}_{[:, :, i]} \in
    \mathbb{R}^{L^{\prime}/2 \times J^\prime}
\end{equation}

In $\textbf{P}_e$, the masking criterion is based on global temporal correlations of the skeleton sequence.
Similar to $\textbf{P}_o$, we utilize the global correlation matrix ($\textbf{GCM}\in \mathbb{R}^{J^{\prime} \times L^{\prime}/2 \times L^{\prime}/2}$) to represent the correlation values of the same node in different frames.
Here, we compare the model's performance under two different strategies:
\begin{itemize}
    \item \textbf{Strategy 1}: Based on the features already embedded in hyperbolic space. The correlation matrix is computed using the cosine of hyperbolic distances between joints. The value at the $i_{th}$ row and $j_{th}$ column in the \textbf{GCM} of the $k_{th}$ joint can be represented as:
    \begin{equation}\label{eq：inner}
    \textbf{GCM}^{ij}_{k} = -\cosh(d_{\mathbb{P}}(u^i, u^j)) + 1
    \end{equation}
    where $u^i$ and $u^j$ are representations of the $k_{th}$ joint in different frames, the correlation matrix is calculated independently for each joint.
    Adding 1 normalizes the hyperbolic cosine distance into a relative distance measure, ensuring that the values fall within a specific range, typically between 0 and 1.
    
    \item \textbf{Strategy 2}: Based on the features after positional embedding but before mapping to hyperbolic space. The even-indexed portion of the features is selected, and the $\textbf{GCM}^{ij}_{k}$ is directly calculated using the transformer based on Euclidean space:
    \begin{equation}\label{eq:trans}
    \mathbf{GCM}^{ij}_{k}=Softmax\left(\frac{\psi\left(\mathbf{E^i_{k}}\right) \phi\left(\mathbf{E^j_{k}}\right)^T}{\sqrt{d_{c}}}\right)
    \end{equation}
    where $\psi(\cdot)$ and $\phi(\cdot)$ represent linear connection layer or one-dimensional convolution layer.
    $\sqrt{d_{c}}$ is a constant designed to keep the gradient value of the model stable during the training process, usually taking the number of channel dimensions of \textbf{E}, $\textbf{E}^i_{k}$ represents the $k_{th}$ joint in the $i_{th}$ frame of the \textbf{E}.
    
\end{itemize}

After obtaining the \textbf{GCM},  we sum the values in its last dimension and get $\textbf{T}_{C}$, which represents the correlations of the same joint across different frames and is also the criteria for the mask of $\textbf{P}_{e}$.

 \begin{equation}\label{eq:GC}
    \mathbf{T}_{C} = \sum_{i=0}^{L^{\prime}/2-1} \textbf{GCM}_{[:, :, i]} \in
    \mathbb{R}^{J^{\prime} \times L^\prime/2}
\end{equation}

\noindent\textbf{Non-masked Joint Extraction:}
Here, both \(\mathbf{S}_{H}\) and \(\mathbf{T}_{C}\) are further reshaped into \(\mathbf{I}_{S}\) and $\mathbf{I}_{T} \in \mathbb{R}^{(L^{\prime} \times J^\prime)/2}$. To ensure consistent indexing when concatenating the two parts, \(\mathbf{T}_{C}\) is transposed before reshaping. The index of the $i_{th}$ joint in the $k_{th}$ frame in both \(\mathbf{P}_e\) and \(\mathbf{P}_o\) is given by \(k \times J^\prime + i\).

Sec.\ref{sec: intro} emphasizes that introducing a degree of randomness during masking is essential for ensuring the model's generalization capability. Consequently, the Gumbel-Max is applied to randomize the reshaped \(\mathbf{I}_{S}\) and \(\mathbf{I}_{T}\), enabling efficient probability-guided mask index sampling:
\begin{equation}\label{eq: gumble}
\begin{split}
& \mathbf{\pi_S}=\mathrm{Softmax}\left(\frac{\textbf{I}_{S}/\mathrm{max}\left(\textbf{I}_{S}\right)}{\tau}\right), \\
&g=-\log_{}{\left(-\log_{}{\eta}\right)},\ \ \ \eta\sim U\left(0,1\right),\\
&idx^{umask}_s=\mathrm{argsort}\left(\log_{}{\mathbf{\pi_S}}+g\right)\left[l-M :\right],
\end{split}
\end{equation}
where $\tau$ is a temperature hyperparameter which controls the trade-off between randomness and determinism.
$\eta$ is random noise sampled from a uniform distribution between 0 and 1.
The obtained $idx^{umask}_s$ indicates which joints are unmasked.
Joints with finer detail features are masked, while \(\mathbf{P}_e\) and \(\mathbf{P}_o\) correspond to joints that are lower in the hierarchy and more correlated, forcing the model to learn these peripheral features.
$l=T'/2 \times J'$,
$M$ represents the number of unmasked joints related to the masking rate $r_{m}$.
\(\mathbf{I}_{T}\) undergoes the same operations to obtain ${idx^{umask}_t}$.

Based on \({idx^{umask}_t}\) and \({idx^{umask}_s}\), the non-masked joints can be extracted, and the corresponding binary mask matrices in \(\mathbf{P}_o\) and \(\mathbf{P}_e\) can be obtained, where 1 represents the index of a non-masked node, and 0 indicates the opposite.

Next, the mask matrices, non-masked joint features, and unmask indices are concatenated separately. 
\begin{equation}\label{eq: cat joint}
{\textbf{E}_{um}}= Concat({\mathbf{P}_o \odot {idx^{umask}_t}}, \mathbf{P}_e \odot {idx^{umask}_s}))
\end{equation}
where $\odot$ represents extraction operation.
Note that after the odd-even cross-grouping, the range of joint indices in both parts is \([0:l]\). Therefore, when concatenating the unmask indices, all indices in \(\mathbf{P}_e\) must have \(l\) added to them:
\begin{equation}\label{eq: cat index}
{idx^{umask}}= Concat({idx^{umask}_t}, {idx^{umask}_s} + l))
\end{equation}

\noindent\textbf{Encoder:}
$\textbf{E}_{um} \in\mathbb{R}^{2M \times C^\prime}$ is used as the input of Encoder.
$L_{e}$ layers of vanilla transformer \cite{r38} blocks are applied to extract latent representations. Each block comprises a multi-head self-attention (MSA) module and a feed-forward network (FFN) module. Residual connectivity is applied within each module, followed by layer normalization (LN):
\begin{equation}
\begin{split}
&\textbf{E}_{0}=E_{um},\\
&\textbf{E}_{l}'=\mathrm{MSA}\left(\mathrm{LN}\left(\textbf{E}_{l-1}\right)\right)+\textbf{E}_{l-1},\ \ \ l\in 1,\dots L_{e}\\
&\textbf{E}_{l}=\mathrm{MLP}\left(\mathrm{LN}\left(\textbf{X}_{l}'\right)\right)+\textbf{X}_{l}',\ \ \ \ \ \ \ \  l\in 1,\dots L_{e}\\
&\textbf{E}_{e}=\mathrm{LN}\left(\textbf{X}_{L_e}\right),
\end{split}
\end{equation}
where $\textbf{E}_{e}$ will serve as the input of the Decoder for reconstruction and directly output to guide the model's cross contrast loss.

\noindent\textbf{Decoder:} 
\(\textbf{E}_{e}\) contains the latent representations of the visible encoded tokens and learnable mask tokens are inserted into it at positions specified by \(idx^{umask}\), resulting in \(\textbf{E}_{d} \in \mathbb{R}^{2l \times C^\prime}\).
Similar to the encoder,
the decoder employs $L_{d}$ layers of transformer blocks for masked reconstruction:
\begin{equation}
\begin{split}
&\textbf{D}_{0}=\textbf{E}_d+\textbf{P}_{s}+\textbf{P}_{t},\\
&\textbf{D}_{l}'=\mathrm{MSA}\left(\mathrm{LN}\left(\textbf{D}_{l-1}\right)\right)+\textbf{D}_{l-1},\ \ \ l\in 1,\dots L_{d}\\
&\textbf{D}_{l}=\mathrm{MLP}\left(\mathrm{LN}\left(\textbf{D}_{l}'\right)\right)+\textbf{D}_{l}',\ \ \ \ \ \ \ \ l\in 1,\dots L_{d}\\
&\textbf{D}_{d}=\mathrm{LN}\left(\textbf{D}_{L_{d}}\right),
\end{split}
\end{equation}
where \(\mathbf{D}_{d}\) acts as a predictor for target reconstruction. \(\mathbf{P}_s\) and \(\mathbf{P}_t\) represent the spatial and temporal embeddings, respectively, as described in Sec.\ref{sec: pe}.

\subsection{Cross Contrast Loss $\&$ Reconstruction Loss}

The loss function in this paper consists of two components: reconstruction loss, which guides the model in capturing intra-sample details, and cross-contrastive loss, which helps it learn inter-sample differences.

For reconstruction loss, the prediction target is the motion information derived from the first-order difference of the original skeleton sequence with a step size of 1:
\begin{equation}
\mathbf{X}^{target}_i = \mathbf{X}^\prime_{i+1} - \mathbf{X}^\prime_i
\end{equation}
where $i$ $\in 0, 1, ..., L-2$, and the last frame is padded with “0”.
\textbf{X} refers to the original sequence after spatial pruning as described in Eq.\ref{eq:sp}.

To align the dimensions of the prediction target with the reconstruction features, we stack information from \( r \) consecutive frames of \(\mathbf{X}^{target}\) and use a fully connected layer to reduce the feature dimension of \(\mathbf{D}_d\) to \(3r\).
\begin{equation}
\begin{aligned}
& \textbf{X}^{\text {target}}=\| \textbf{X}_{\text {ir:(i+1)r-1 }}^{\text {target}}, i=0,1,...,\frac{L}{r}-1 \\
& \textbf{X}^{\text {pred}}=f_c\left(\textbf{D}_d\right)
\end{aligned}
\end{equation}
where $\|$ is a stacking operation, and $r$ is the same as the convolution kernel size in Eq.\ref{eq: tp}.
$f_c$ represents fully connected layer.
Since \(\mathbf{D}_d\) is embedded in hyperbolic space, we similarly use Eq.\ref{eq:exp} and Eq.\ref{eq:poin} to embed the $\textbf{X}^{\text {target}}$ into the Poincaré ball model.

We utilize the mean squared error (MSE) between the predicted result $\textbf{X}^{\text {pred}}$ and the reconstruction target $\textbf{X}^{\text {target}}$ for masked joints:
\begin{equation}
\mathcal{L}_{\mathrm{r}}=\frac{1}{2M}\sum_{i\notin idx^{umask}}\left\|\textbf{X}_{i}^{{\text {pred}}} - \textbf{X}_{i}^{\text {target}}\right\|_{2}^{2},
\end{equation}
where \( M \) is the same as the number of unmasked joints in Eq.~\ref{eq: gumble}.

For the cross-contrastive loss, the masked features $\textbf{E}_e$ will split again, i.e., divided into the original even and odd components. Since both are subsets of the complete sequence, they exhibit high similarity at the instance-level representation.

Therefore, we first utilize average pooling to obtain the instance-level representation of each component:
\begin{equation}
\textbf{E}_e^o, \textbf{E}_e^e, \textbf{E}_e^c=\text{ Meanpooling}\left( \textbf{E}_e^{\text {odd}}, \textbf{E}_e^{\text{even}}, \textbf{E}_e\right) \in 
\mathbb{R}^{N \times C^\prime}
\end{equation}
where $\textbf{E}_e^{\text {odd}}$,$\textbf{E}_e^{\text{even}}$ represents the odd and even components of $\textbf{E}_e$,respectively, and $N$ is the batch size during training.
Then, the pairwise contrastive losses between the three components (odd, even, and complete) within the same batch are computed, and we sum these losses as the final cross contrast loss:
\begin{equation}
\begin{aligned}
\mathcal{L}_{c^2}= & \left\|\left(\textbf{E}_e^o \cdot \textbf{E}_e^{c T}\right)-\textbf{I}_N\right\|_2^2 + \\
& \left\|\left(\textbf{E}_e^e \cdot \textbf{E}_e^{c T}\right)-\textbf{I}_N\right\|_2^2 + \\
& \left\|\left(\textbf{E}_e^e \cdot \textbf{E}_e^{e T}\right)-\textbf{I}_N\right\|_2^2
\end{aligned}
\end{equation}
where $T$ denotes the transpose operation, and $\textbf{I}_N$ represents the identity matrix of dimension $N$. 

Finally, the total loss is expressed as follows:
\begin{equation}
\mathcal{L}=\mathcal{L}_{r}+\mu\mathcal{L}_{c^2}
\end{equation}
where $\mu$ is a hyperparameter used to weigh these two losses.

\section{Experiments}
We conducted extensive experiments to evaluate whether
our proposed framework can learn effective self-supervised feature representations for the task of skeleton action recognition.
To this end, we evaluated the framework under various experimental settings, including unsupervised, semi-supervised, supervised learning, and transfer learning. These experiments were performed on three publicly available datasets: NTU RGB+D \cite{r34}, NTU RGB+D 120 \cite{r35}, and PKU-MMD \cite{r36}.

\begin{table*}[!ht]
\caption{Performance comparison on the NTU-60, NTU-120, and PKU-MMD datasets under the linear evaluation protocol.}\label{tab:table1}
\centering
\begin{tabular}{cccccccccc}
\hline
\multirow{2}*{Method}&\multirow{2}*{Journal\&Year}&\multirow{2}*{Input Stream}&\multicolumn{2}{c}{NTU-60}&\multicolumn{2}{c}{NTU-120}&PKU-I&PKU-II\\
&&&X-sub&X-view&X-sub&X-set&Phase I&Phase II\\
\hline
AimCLR\cite{r2}&AAAI'22&Joint+Motion+Bone&78.9&83.8&68.2&68.8&87.4&39.5\\
SkeAttnCLR\cite{r51}&IJCAI'23&Joint+Motion+Bone&82.0&86.5&77.1&80.0&89.5&55.5\\
ActCLR\cite{r40}&CVPR'23&Joint+Motion+Bone&84.3&88.8&74.3&75.7&90.0&55.9\\
PCM$^{3}$\cite{r41}&MM'23&Joint+Motion+Bone&87.4&\textbf{93.1}&80.0&81.2&-&\textbf{58.2}\\
Colorization\cite{r14}&TPAMI'23&Joint+Motion+Bone&79.1&87.2&69.2&70.8&89.2&49.8\\
Skeleton-logoCLR\cite{r42}&TCSVT'24&Joint+Motion+Bone&86.1&89.8&79.8&80.1&92.2&57.7 \\
\textbf{HA-CM (ours)}&&Joint+Motion+Bone&\textbf{88.0}&92.5&\textbf{82.2}&\textbf{82.5}&\textbf{92.6}&55.0
\\

\hline
P\&C\cite{r9}&CVPR'20&Joint&50.7&76.3&42.7&41.7&59.9&25.5\\
CMD\cite{r6}&ECCV'22&Joint&79.4&86.9&70.3&71.5&-&43.0\\
HaLP\cite{r45}&CVPR'23&Joint&79.7&86.8&71.1&72.2&43.5\\
HiCo\cite{r7}&AAAI'23&Joint&81.1&88.6&72.8&74.1&89.3&49.4\\
PCM$^{3}$\cite{r41}&MM'23&Joint&83.9&90.4&76.5&77.5&-&51.5\\
H$^2$E\cite{r23}&TIP'23&Joint&78.7&82.3&-&-&88.5&51.7\\
Skeleton-logoCLR\cite{r42}&TCSVT'24&Joint&82.4&87.2&72.8&73.5&90.8&\textbf{54.7}\\
SCD-Net\cite{rIN9}&AAAI'24&Joint&\textbf{86.6}&\textbf{91.7}&76.9&80.1&91.9&54.0\\

\hline
\rowcolor{gray!10}\textit{MAE-like Methods:}&&&&&&&&\\
SkeletonMAE\cite{r46}&ICME'21&Joint&74.8&77.7&72.5&73.5&82.8&36.1\\
MAMP\cite{r12}&ICCV'23&Joint&84.9&89.1&78.6&79.1&92.2&53.8\\
MMFR\cite{r48}&TCSVT'24&Joint&84.2&89.5&77.1&78.8&\textbf{92.4}&54.4\\
\textbf{HA-CM (ours)}&&Joint&86.3&91.2&\textbf{78.9}&\textbf{80.2}&91.6&50.9\\
\hline
\end{tabular}
\end{table*}

\subsection{Datasets}
\noindent\textbf{NTU-RGB+D 60:} NTU-RGB+D 60 (NTU60) contains 56,880 3D skeleton sequences across 60 types of actions, and all actions are performed by 40 subjects.
It uses two division criteria when dividing the training and test sets.
1) X-Sub: The training set and test set are divided according to the person ID, with the test set containing 16,487 sequences.
2) X-View: The training set and the test set are divided according to the camera, with the test set (camera 1) containing 18,932 sequences.

\noindent\textbf{NTU-RGB+D 120:} NTU-RGB+D 120 (NTU120) extends 60 categories based on NTU-RGB+D 60, and the number of total skeleton sequences and subjects are also increased to 114,480 and 106, respectively.
It recommends two subsets similar to NTU-60: X-Sub and X-Set. In X-Sub, training data is derived from 50$\%$ of the subjects, while the remaining 50$\%$ are used for testing. In X-Set, training data comes from samples with even setup IDs, and testing data is drawn from samples with odd setup IDs.

\noindent\textbf{PKU-MMD:} PKU-MMD  consists of 1,076 uncut video sequences captured from 3 different camera viewpoints, involving 66 participants, including 51 annotated action categories. Following the approach in \cite{r37}, we segment the dataset into two phases: PKU-MMD I (PKU-I) and PKU-MMD II (PKU-II). In PKU-I, the training set contains 18,841 samples, while the test set has 2,704 samples. PKU-II includes 5,332 samples for training and 1,613 samples for testing.

\subsection{Implementation Details}
\noindent\textbf{Network Architecture:} The proposed $\textbf{HA-CM}$ in this paper is built upon an Encoder-Decoder framework, where both the Encoder and Decoder utilize the vanilla Transformer architecture. The encoder comprises $L_{e}$ = 8 identical building blocks, while the decoder comprises $L_{d}$ = 3 blocks.
Each block features an embedding dimension of 256, the head number of the multi-head self-attention module is set to 8, and the hidden dimension is 1024. 

\noindent\textbf{Data Processing Details:} 
During pre-training, we randomly cropped with a certain proportion $p$ sampled from [0.5,1] during the training phase. The number of frames for the cropped clip is fixed to $L$ = 72 by bilinear interpolation. For testing, $p$ is fixed to 0.9. Since $L$ influences the amount of information in the sequence and consequently affects the model's performance, we conducted ablation experiments to explore the impact of varying $L$.

\noindent\textbf{Pre-training Details:} During pre-training, the masking ratio of the encoder input tokens is set to 90\%. We use the AdamW optimizer with a weight decay of 0.05 and betas of (0.9, 0.95). 
The network was trained for 400 epochs, with the learning rate linearly increasing to 1e-3 from 0 during the first 20 warm-up epochs, followed by a decay to 5e-4 according to a cosine schedule. Our model is implemented using the PyTorch framework, and the experiments are conducted on two NVIDIA RTX 4090 GPUs.

\subsection{Comparison with State-of-the-art Methods}
\begin{table*}[!ht]
\caption{Performance comparison on the NTU-60 and NTU-120 datasets under the fine-tuning evaluation protocol.\label{tab:table2}}
\centering
\begin{tabular}{cccccccccc}
\hline
\multirow{2}*{Method}&\multirow{2}*{Journal\&Year}&\multirow{2}*{Input Stream}&\multirow{2}*{Backbone}&\multicolumn{2}{c}{NTU-60}&\multicolumn{2}{c}{NTU-120}\\
&&&&X-sub&X-view&X-sub&X-set\\
\hline
CrosSCLR\cite{r5}&CVPR'21&Joint+Motion+Bone&ST-GCN&86.2&92.5&80.5&80.4\\
AimCLR\cite{r2}&AAAI'22&Joint+Motion+Bone&ST-GCN&86.9&92.8&80.1&80.9\\

Colorization\cite{r14}&TPAMI'23&Joint+Motion+Bone&DGCNN&89.1&95.9&81.2&82.4\\
ActCLR\cite{r40}&CVPR'23&Joint+Motion+Bone&ST-GCN&88.2&93.9&82.1&84.6\\
HYSP\cite{r22}&ICLR'23&Joint+Motion+Bone&ST-GCN&89.1&95.2&84.5&86.3\\
Skeleton-logoCLR\cite{r42}&TCSVT'24&Joint+Motion+Bone&ST-GCN&89.4&94.3&84.6&85.7\\

\hline
SkeletonMAE\cite{r46}&ICME'21&Joint&STTFormer&86.6&92.9&76.8&79.1\\
MotionBERT\cite{r50}&ICCV'23&Joint&DSTformer&93.0&97.2&-&-\\
MAMP\cite{r12}&ICCV'23&Joint&Transformer&\textbf{93.1}&97.5&\textbf{90.0}&\textbf{91.3}\\
MMFR\cite{r48}&TCSVT'24&Joint&Transformer&91.9&96.5&87.4&90.4\\

\textbf{HA-CM (ours)}&&Joint&Transformer&92.4&\textbf{97.7}&89.1&90.5\\
\hline
\end{tabular}
\end{table*}

\noindent\textbf{Linear Evaluation:} 
we retain the Encoder, prior refinement, and hyperbolic mapping components. 
The weights of the pre-trained backbone are fixed, and a post-attached linear classifier is added for supervised classification. 
The network is trained for 100 epochs using SGD with a batch size of 128. The learning rate starts at 0.1 and decays to 0 following a cosine schedule. The results are illustrated in Tab.\ref{tab:table1}, where we categorize the methods in the table into three groups:3-streams networks,
non-masking reconstruction, and masking reconstruction Methods.
\begin{itemize}
    \item The methods within the 3-stream networks all belong to contrastive learning paradigm. To our knowledge, HA-CM is the first work to demonstrate 3-stream results under a non-contrastive learning approach. HA-CM's results are obtained from a supervised ensemble that integrates joint, velocity, and bone, weighted equally. HA-CM outperforms in four out of six metrics across the three datasets, and demonstrating strong competitiveness in the remaining two.

    \item The leading method in non-masking reconstruction, SCD-Net\cite{rIN9}, outperforms HA-CM by 0.3\% and 0.5\% on four metrics in the NTU60 and PKU datasets, respectively. However, in the larger NTU120 dataset, HA-CM shows advantages of 2.0\% and 0.1\%, highlighting its superior performance in more complex scenarios.

    \item In the masking reconstruction paradigm, our method outperforms the baseline MAMP by 1.4\%, 2.1\%, 0.3\%, and 1.1\%  across four metrics in the NTU60 and NTU120 datasets. However, HA-CM shows a decline in performance on the PKU dataset, likely due to its smaller size, which makes the model more susceptible to amplifying noise effects in high-dimensional hyperbolic space, leading to misjudgments.
\end{itemize}

\begin{table}[!t]
\caption{Performance comparison on the NTU-60 dataset under the semi-supervised evaluation protocol.\label{tab:table3}}
\centering
\begin{tabular}{cccccccccc}
\hline
\multirow{3}*{Method}&\multirow{3}*{Journal\&Year}&\multicolumn{4}{c}{NTU-60}\\
&&\multicolumn{2}{c}{X-sub}&\multicolumn{2}{c}{X-view}\\
&&(1\%)&(10\%)&(1\%)&(10\%)\\
\hline
3s-CrosSCLR\cite{r5}&CVPR'21&51.1&74.4&50.0&77.8\\
SkeletonMAE\cite{r46}&ICME'21&54.4&80.6&54.6&83.5\\
3s-AimCLR\cite{r40}&AAAI'22&54.8&78.2&54.3&81.6\\
3s-CMD\cite{r6}&ECCV'22&55.6&79.0&55.5&82.4\\
3s-HYSP\cite{r22}&ICLR'23&-&80.5&-&85.4\\
Colorization\cite{r14}&TPAMI'23&52.3&76.5&53.1&81.3\\
3s-SkeAttnCLR\cite{r51}&IJCAI'23&59.6&81.5&59.2&83.8\\
SCD-Net\cite{rIN9}&AAAI'24&69.1&82.2&66.8&85.8\\
MMFR\cite{r48}&TCSVT'24&65.0&87.0&71.3&91.0\\
\hline
MAMP\cite{r12}&ICCV'23&66.0&88.0&68.7&91.5\\
\textbf{HA-CM (ours)}&&\textbf{69.3}&\textbf{88.2}&\textbf{74.0}&\textbf{92.2}\\
\hline
\end{tabular}
\end{table}

\noindent\textbf{Fine-tuning Evaluation:} we apply an MLP head after the pre-trained student encoder and fully fine-tuned the retained framework for 100 epochs with a batch size of 48. We utilize the AdamW optimizer with a weight decay of 0.05. The learning rate is set to 0 and linearly increased to 3e-4 for the first 5 warm-up epochs, then decreased to 1e-5 following a cosine decay schedule. 

As presented in Tab. \ref{tab:table2},  masking reconstruction methods consistently outperform contrastive learning approaches, with even single-stream masking surpassing three-stream contrastive methods.
This advantage likely arises from masking's ability to suppress input noise, enhancing model accuracy by focusing on effective sample information. In contrast, contrastive learning is vulnerable to noise, which can be misinterpreted as valid features, degrading learning quality.
Compared to the baseline MAMP\cite{r12}, HA-CM shows variations of -0.7\%, +0.2\%, -0.9\%, and -0.8\% across four metrics. This contradiction with the results from linear evaluation in downstream tasks may be due to the simplistic MLP head being unable to fully leverage the complex features learned by HA-CM.

\begin{table}[!t]
\caption{Performance comparison under the transfer learning protocol.\label{tab:table4}}
\centering
\begin{tabular}{cccccccccc}
\hline
\multirow{2}*{Method}&\multirow{2}*{Journal\&Year}&\multicolumn{2}{c}{To PKU-II}\\
&&NTU-60&NTU-120\\
\hline
LongT GAN\cite{r8}&AAAI'18&44.8&-\\
SkeletonMAE\cite{r46}&ICME'21&58.4&61.0\\
CMD\cite{r6}&ECCV'22&56.0&57.0\\
SCD-Net\cite{rIN9}&AAAI'24&67.5&-\\
MMFR\cite{r48}&TCSVT'24&68.7&69.7\\
\hline
MAMP\cite{r12}&ICCV'23&70.6&\textbf{73.2}\\
\textbf{HA-CM (ours)}&&\textbf{71.1}&72.3\\
\hline
\end{tabular}
\end{table}

\noindent\textbf{Semi-supervised Evaluation:} The post-attached classification layer and the pre-trained student encoder are fine-tuned together, utilizing only 1\% and 10\% of the training data to align with the fine-tuning evaluation protocol. Importantly, to account for randomness in data selection, the reported results represent the average of five runs. 

As shown in Tab. \ref{tab:table3},
HA-CM demonstrates state-of-the-art performance, surpassing the baseline MAMP by 3.3\%, 0.2\%, 5.3\%, and 0.7\% across four metrics. This indicates that HA-CM is more robust in handling noisy or partially labeled data, reducing reliance on annotation quality while effectively capturing the underlying features and patterns in the data. Such capabilities enhance its adaptability to complex tasks, aligning well with the strengths of self-supervised learning models.

\noindent\textbf{Transfer Learning Evaluation:}
The pre-training of HA-CM is pre-trained on the source datasets, NTU-60 (X-Sub) and NTU-120 (X-Sub), before being fine-tuned on the target dataset, PKU-MMD II, in our experiments.

As shown in Tab.\ref{tab:table4}, when transferring the model trained on the NTU60 dataset to PKU, HA-CM achieved the best performance, demonstrating its strong transferability. However, when transferring the model trained on the NTU120 dataset to PKU, the results decreased by 0.9\% compared to MAMP, likely due to the smaller sample size of PKU, which may not adequately represent the diversity of the NTU120 dataset. Nonetheless, HA-CM still demonstrates excellent feature transferability and remains a suboptimal result.

\begin{figure}[!t]
\centering
\includegraphics[width=7.5cm]{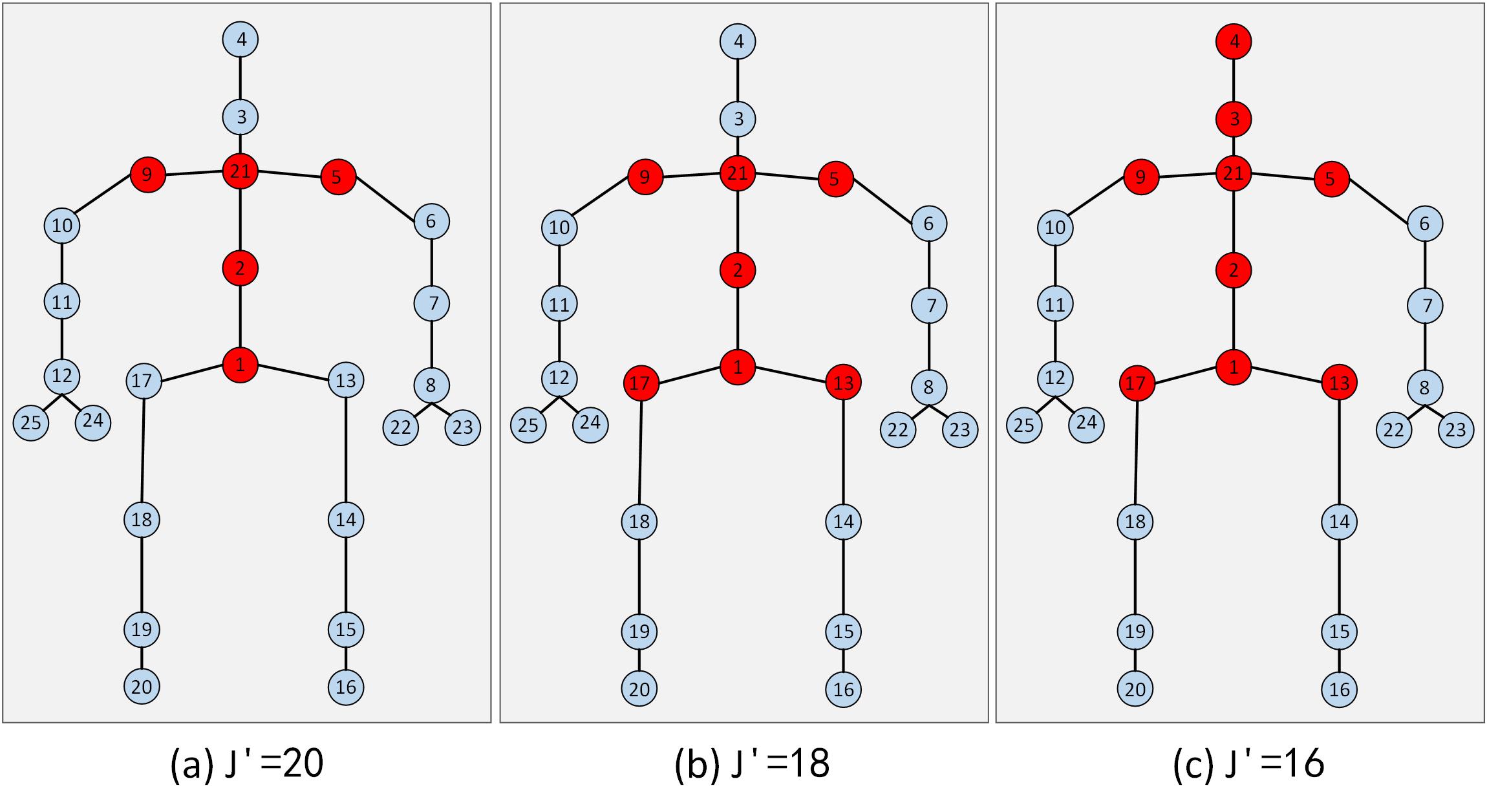}
\caption{Joint removal in the spatial pruning module using the NTU dataset. Blue joints are retained, red joints are removed, and $J^\prime$ denotes the number of joints after pruning.}
\label{fig4}
\end{figure}

\begin{table}[!ht]
    \centering
    \caption{Impact of Sequence Redundancy on Model Performance and Training Efficiency}
    \label{tab5}
    \begin{tabular}{c|ccc|cc}
 \hline
    \textbf{Configurations} & $J'$ & $L$ & $r$ & \textbf{Acc(\%)} & \textbf{Time(Hours)} \\ 
    \hline
    Original & 25 & 120 & 4 & 84.9 & - \\ 
    \hline
    Reproduction & 25 & 120 & 4 & 84.7 & 14.3 \\ 
    \hline
    \multirow{3}*{Spatial Pruning} & 20 & 120 & 4 & 84.8 & 9.9 \\ 
    & \textbf{18} & \textbf{120} & \textbf{4} & \textbf{85.1} & \textbf{8.6} \\ 
    & 16 & 120 & 4 & 81.5 & 7.6 \\ 
    \hline
    \multirow{3}*{Temporal Clip} & 25 & 92 & 4 & 84.3 & 9.1 \\ 
    & 25 & 72 & 4 & 83.8 & 7.8 \\ 
    & 25 & 64 & 4 & 82.0 & 6.5 \\ 
    \hline
    \multirow{3}*{Pooling Size} & \textbf{25} & \textbf{120} & \textbf{3} & \textbf{84.9} & \textbf{19.2} \\ 
    & 25 & 120 & 5 & 84.4 & 11.5 \\ 
    & 25 & 120 & 6 & 84.2 & 9.5 \\ 
    \hline
    \multirow{4}*{Ablation} & \textbf{18} & \textbf{72} & \textbf{3} & \textbf{85.2} & \textbf{6.6} \\ 
    & 18 & 90 & 3 & 85.0 & 8.5   \\ 
    & 25 & 72 & 3 & 84.9 & 9.5 \\ 
    & 18 & 72 & 4 & 85.0 & 5.1 \\ 
 \hline
    \end{tabular}
\end{table}

\begin{table}[!t]
\caption{The Necessity of Maintaining Randomness in Masking Process, Temperature Parameter ($\tau$ = 0.8) in Gumble max.}\label{tab6}
\centering
\begin{tabular}{cccc}
\hline
Method& Gumble Max&Acc(\%)\\
\hline
\multirow{2}*{MAMP}&\checkmark&84.9\\
& $\times$ &66.7 \\
\hline
\multirow{2}*{HA-CM} &\checkmark&\textbf{85.9}\\
&$\times$ & 68.1 \\
\hline
\end{tabular}
\end{table}

\begin{figure}[!t]
\centering
\includegraphics[width=8cm]{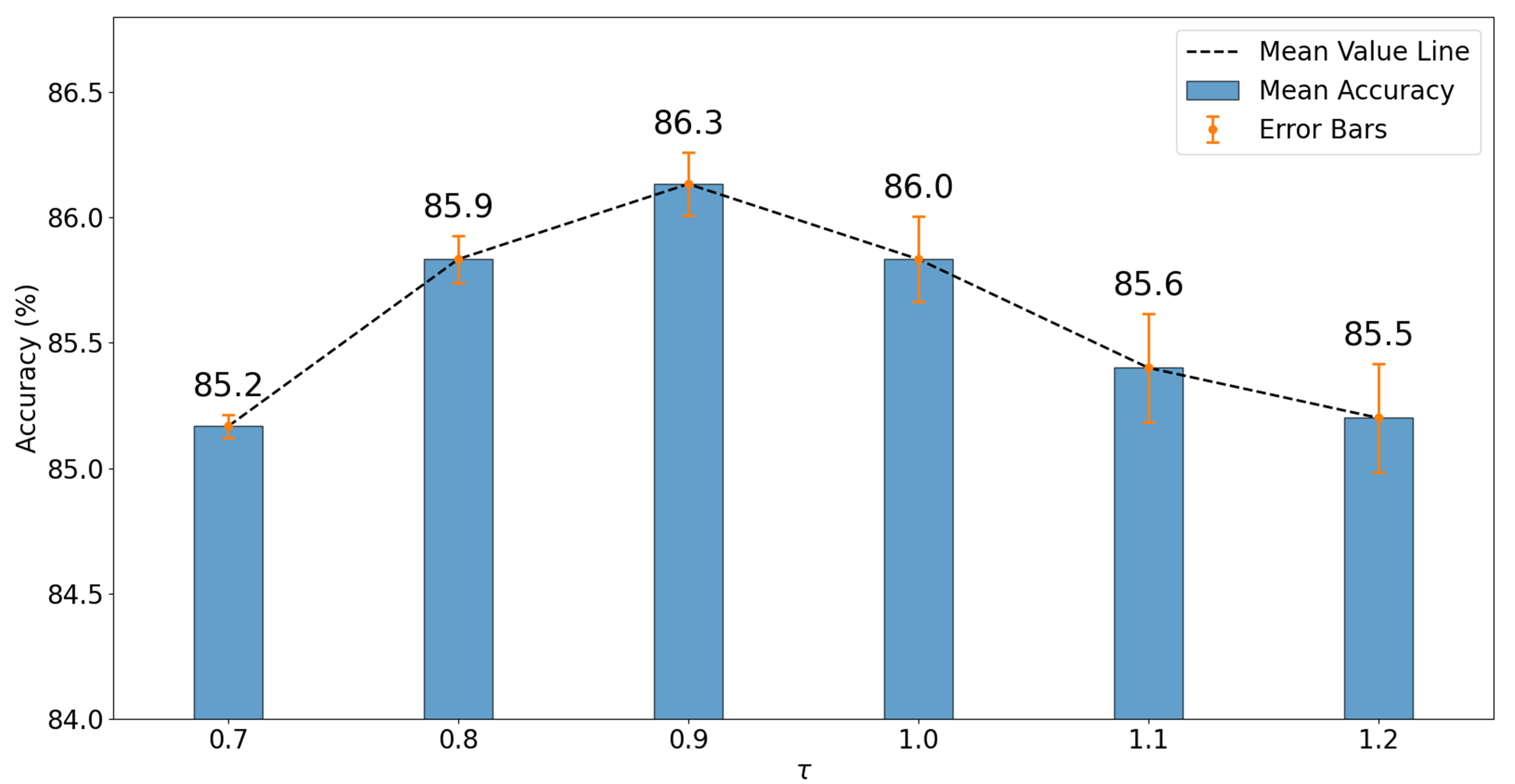}
\caption{Model Performance vs Temperature Coefficient ($\tau$). The error bars obtained from training each parameter three runs.}
\label{fig5}
\end{figure}

\subsection{Ablation Study}
We conducted extensive ablation studies on the NTU-60 (sub) dataset to analyze the proposed HA-CM. Unless otherwise specified, we report the results under the linear evaluation protocol.

\noindent\textbf{Prior Refinement:}
The proposed HA-CM utilizes a mask reconstruction paradigm. Unlike contrastive learning methods, which emphasize input richness, this approach focuses on reconstructing masked portions, making edge features (core information) more critical. Excessive information can impede learning by diverting attention from the reconstruction task. This insight guided the design of our prior refinement module.

Before constructing HA-CM, we investigated the impact of skeleton sequence redundancy on model performance based on the baseline MAMP \cite{r12}. Specifically, We explored how varying the number of spatial joints, sequence length, and convolution kernel size in temporal pooling affects the model's accuracy and training time. Fig.\ref{fig4} illustrates the schematics for different numbers of spatial joints $J^\prime$ after pruning, and detailed results are presented in Tab.\ref{tab5}. We have three observations as follows:

\begin{itemize}
    \item In Spatial Pruning, pruning joints from the torso significantly reduces training time while maintaining or even improving model accuracy. This indicates their lower contribution to the mask reconstruction task. Conversely, removing head joints causes a sharp accuracy drop, underscoring their critical role in capturing essential features for model performance. 
    \item  With a fixed pooling size, shorter sequences result in less comprehensive feature extraction. Conversely, a smaller pooling size with a fixed sequence length increases the token count, potentially exacerbating redundancy issues and extending training time. Therefore, finding the right balance between sequence length (for comprehensive feature extraction) and pooling size (for feature resolution) is crucial.
    \item The best performance is achieved with 18 spatial joints, 72 frames, and a kernel size of 3, resulting in an accuracy of 85.2$\%$ and a training time of 6.6 hours. This combination optimally balances feature retention and training efficiency, which is why HA-CM adopts these settings.
\end{itemize}

\noindent\textbf{Gumble Max:}
Gumbel-Max transforms the masking criteria from a deterministic numerical approach into a probabilistic distribution within the mask reconstruction paradigm, thereby introducing randomness into the masking process. This randomness ensures that the masked nodes retain a certain degree of high-frequency information, representing the data's critical features. Consequently, the model is better equipped to capture and utilize this high-frequency information during training, enhancing its reconstruction capabilities and overall performance.

Tab.\ref{tab6} illustrates that both models exhibit improved performance with the Gumbel-Max technique, confirming the importance of incorporating randomness into the masking process. The Temperature Parameter $\tau$ = 0.8 used to control randomness in both models is consistent with the settings from MAMP. This naturally led us to investigate how variations in this hyperparameter impact the performance of HA-CM.

\noindent\textbf{Temperature Parameter $\tau$:}
From Eq.\ref{eq: gumble}, it is evident that increasing the temperature parameter $\tau$ smooths the probability distribution in Gumbel-Max sampling, resulting in more uniform mask probabilities across joints and higher randomness. Conversely, a decrease in $\tau$ causes the probability distribution to become more concentrated around the higher values in the mask criteria sequence, thereby reducing randomness.

The error bars in Figure \ref{fig5}, derived from training each parameter three times, corroborate this observation: higher $\tau$ values correspond to greater variability in model performance. 
Notably, HA-CM exhibits optimal performance at $\tau$ = 0.9, where the average accuracy peaks and the maximum value reaches 86.3\%. Consequently, the model with $\tau$ = 0.9 is retained as the most effective configuration and is used for subsequent downstream tasks.

\begin{table}[!ht]
    \centering
    \caption{Ablation study on various components in HA-CM.T1 and T2 represent temporal globality under Strategy 1 and 2, respectively, while S represents the spatial hierarchy for mask criterion calculation.}\label{tab7}
    \begin{tabular}{ccccc|c}
    \hline
        ~ & Mask Structure & \multicolumn{2}{c}{Mask Criteria}  & $\mathcal{L}_{c^2}$ & Acc(\%)  \\ 
        \hline
        \multirow{4}*{Baseline} & \multirow{4}*{Uniform Mask} & \multicolumn{2}{c}{Motion Intensity} & $\times$  & 85.2  \\
    
        ~ & ~ & \multicolumn{2}{c}{T1} & $\times$ & 85.6  \\
    
        ~ & ~ & \multicolumn{2}{c}{T2} & $\times$ & 84.6  \\ 
       
        ~ & ~ & \multicolumn{2}{c}{S} & $\times$  & 84.0  \\ 
        \hline
        \multirow{9}*{HA-CM} & \multirow{9}*{Cross Mask} & Odd & Even & ~ & ~  \\ 
        ~ & ~ & \multirow{2}*{T1} & \multirow{4}*{S} &$\times$ & 86.0  \\ 
        
        ~ & ~ & ~ & ~ & \checkmark & \textbf{86.3}  \\
        ~ & ~ & \multirow{2}*{T2} & ~ & $\times$ & 85.6  \\ 
        
        ~ & ~ & ~ & ~ & \checkmark & 85.8  \\ 
        \cline{3-6}
        ~ & ~ & \multirow{4}*{S} & \multirow{2}*{T1} & $\times$ & 85.2  \\ 
        ~ & ~ & ~ & ~ & \checkmark & 85.9  \\ 

        ~ & ~ & ~ & \multirow{2}*{T2} & $\times$ & 84.9  \\ 
        ~ & ~ & ~ & ~ & \checkmark & 85.5  \\ 
        \hline
    \end{tabular}
\end{table}

\noindent\textbf{Ablation Study on HA-CM.}
After determining the refined skeleton sequence format and the temperature parameter $\tau$, we explore the effectiveness of different components in HA-CM and conduct an ablation study on them, as shown in Tab. \ref{tab7}. Our observations are summarized as follows:
\begin{itemize}
    \item 
    A temporal perspective in a single masking approach enhances sequence feature extraction, as shown by the first three rows, which outperform the fourth row.
    The inner product relationship between joints better represents their interactions, as shown by the first 85.2\% and the second line 85.6\%.
    Using hyperbolic distances to represent joint relationships can degrade model performance. As noted in Sec.\ref{sec: intro}, joints function as a Euclidean chain in the temporal dimension, where their relationships are mainly linear. Mapping these to hyperbolic space may amplify perturbations, negatively affecting model performance.
    \item Advantages of the Cross Mask. The cross-mask framework shows a significant performance improvement compared to the global uniform mask. The average accuracy of the model under cross-mask is significantly higher than that with the global uniform mask.
    For example, in the Uniform mask scenario, the individual accuracies for T2 and S are only 84.6\% and 84.0\%, respectively. However, when these two strategies are combined using cross-masking, the model's accuracy increases to 85.6\%. This demonstrates that the cross-masking approach significantly enhances the model's ability to perceive information.
    \item The presence or absence of $\mathcal{L}{c^2}$ significantly impacts the model's accuracy.  
    In the table, all models that include $\mathcal{L}{c^2}$ outperform those without it, especially when baseline performance is lower. This highlights $\mathcal{L}_{c^2}$'s strong generalization capability and its effective synergy with the cross-masking strategy.
\end{itemize}

\noindent\textbf{Hyperparameter $\mu$.} 
Tab \ref{tab8} explores the hyperparameter $\mu$ that controls the weights of the reconstruction loss and contrastive loss in the NTU60 dataset's sub and view settings. 
The results indicate that when $\mu$ is set to 1, the performance is optimal in both settings, with a decreasing trend observed when $\mu$ is either increased or decreased.

\begin{table}[!t]
\caption{Impact of Weighting Between Masked Reconstruction Loss and Cross-Contrastive Loss on Model Performance}\label{tab8}
\centering
\begin{tabular}{cccccccccc}
\hline
$\mu$&0.5&0.8&1.0&1.2&1.5\\
\hline
X-Sub&85.4&86.0&\textbf{86.3}&85.9&85.4\\
X-View&89.8&90.8&\textbf{91.2}&90.5&90.0\\
\hline
\end{tabular}
\end{table}

\begin{figure}[!t]
\centering
\includegraphics[width=8cm]{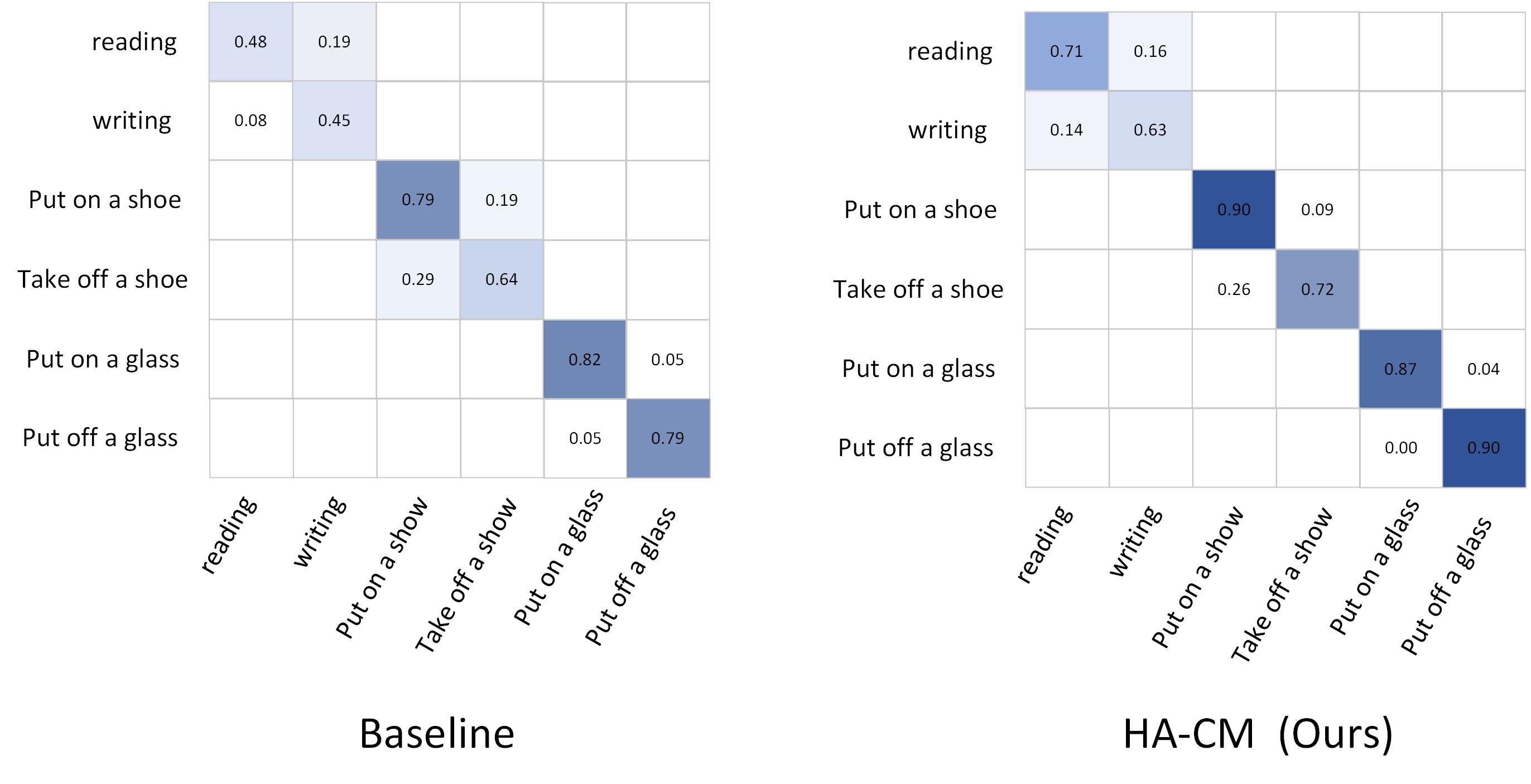}
\caption{Confusion matrix on NTU-60 Xview. The baseline is MAMP.}
\label{fig6}
\end{figure}

\begin{figure}[!t]
\centering
\includegraphics[width=8cm]{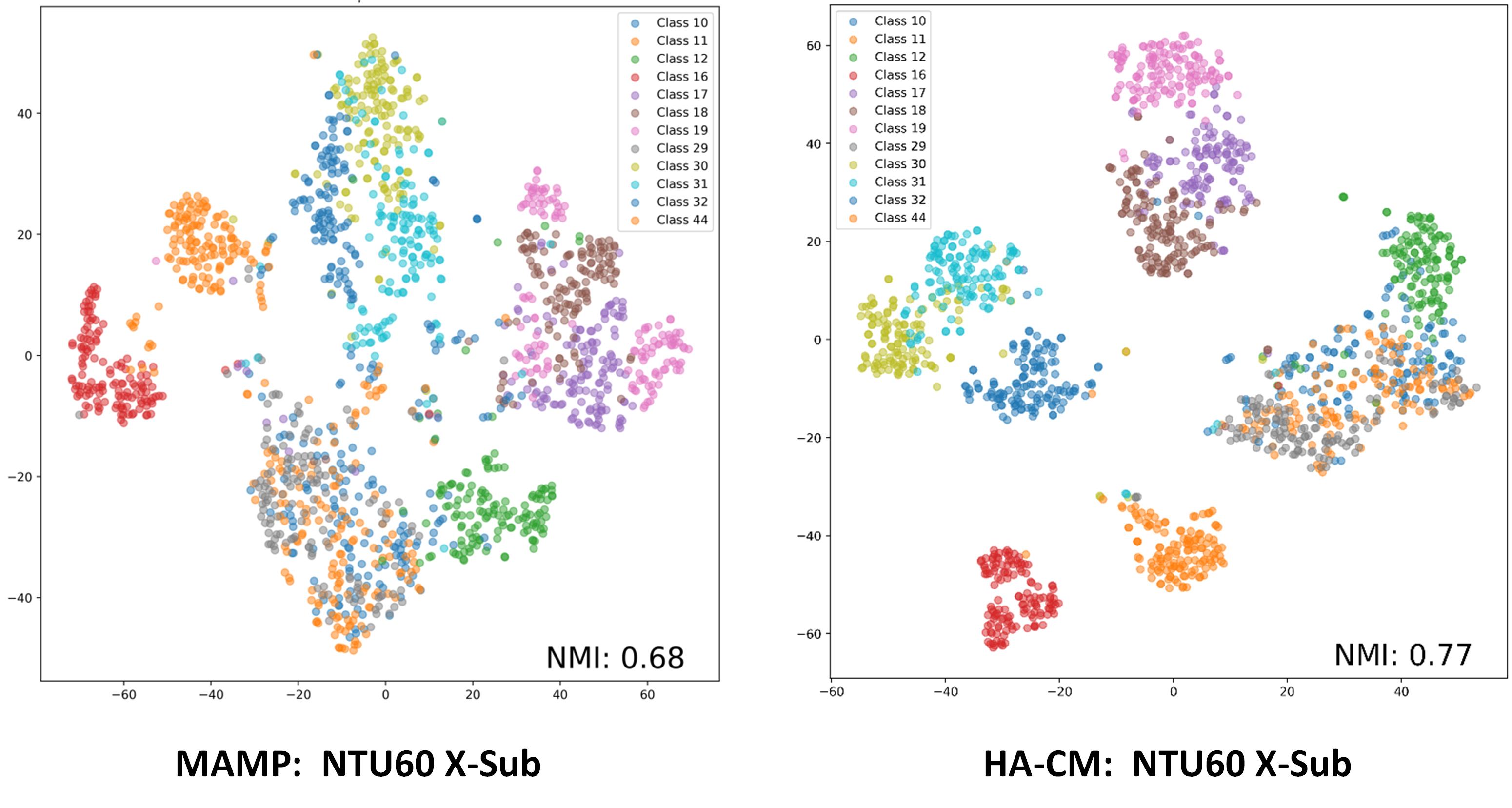}
\caption{The t-SNE visualization on the NTU RGB-D 60 X-Sub. }
\label{fig7}
\end{figure}

\subsection{Qualitative Results} 

 To intuitively demonstrate the superiority of HA-CM in capturing sample details, we conducted visualizations for both MAMP and HA-CM, including a confusion matrix (Fig.\ref{fig6}NTU60 X-View) and a t-SNE plot (Fig.\ref{fig7} NTU60 X-Sub). The confusion matrix features three pairs of highly similar sample classes, while the t-SNE plot expands on this by including three additional pairs. The results demonstrate that HA-CM effectively mitigates inter-class similarity issues, as indicated by the increased diagonal values in the confusion matrix and the reduced misclassification rates for adjacent samples. In the t-SNE plot, normalized mutual information (NMI) measures cluster similarity, with higher values signifying tighter clustering. It shows that HA-CM successfully separates similar samples, highlighting its capability to capture more representative features.


\section{Conclusion}
In this paper, we introduce the hierarchy and attention-guided cross-masking framework (HA-CM), which enhances skeleton sequence analysis by applying masking from both spatial and temporal perspectives, thereby addressing the biases of traditional single masking methods. 
In the temporal flow, we utilize the global attention of joints to replace traditional distance metrics, effectively capturing motion dynamics while overcoming the convergence of distances in high-dimensional space. 
By employing hyperbolic space, HA-CM preserves joint distinctions and maintains the hierarchical structure of high-dimensional skeletons, using joint hierarchy as the masking criterion. 
This work is the first to consider the masking criteria in mask reconstruction from a spatial perspective, leveraging the inherent hierarchical structure of the human skeleton, and it provides a foundation for future masking efforts in the spatial domain, including applications in non-skeletal data.

\bibliographystyle{IEEEtran}
\bibliography{reference}

\end{document}